\documentclass[10pt,twocolumn,letterpaper]{article}

\usepackage{cvpr}      

\newcommand{\datasetName}{CANVAS}
\newcommand{\methodName}{OMGTex}

\definecolor{cvprblue}{rgb}{0.21,0.49,0.74}
\usepackage[pagebackref,breaklinks,colorlinks,allcolors=cvprblue]{hyperref}

\usepackage{multirow}

\def\paperID{} 
\def\confName{CVPR}
\def\confYear{2026}

\title{OMGTex: One-stage Multi-style Facial Texture Reconstruction \\without Geometry Guidance} 

\author{
Zitong Xiao$^{1*}$ \quad
Yuda Qiu$^{1*\dagger}$ \quad
Zisheng Ye$^{1}$ \quad
Xiaoguang Han$^{1,2,3\dagger}$\\
$^{1}$School of Science and Engineering, The Chinese University of Hong Kong, Shenzhen  \\ 
$^{2}$Guangdong Provincial Key Laboratory of Future Networks of Intelligence \\
$^{3}$FNii-Shenzhen \\
{\small $^*$Equal contribution\quad $^\dagger$Corresponding author}
}

\begin{document}
\twocolumn[{%
  \renewcommand\twocolumn[1][]{#1}%
\maketitle
\begin{center}
  \newcommand{\teaserwidth}{\textwidth}
  \vspace{-0.15in}
  \centerline{
    \includegraphics[width=1.0\teaserwidth]{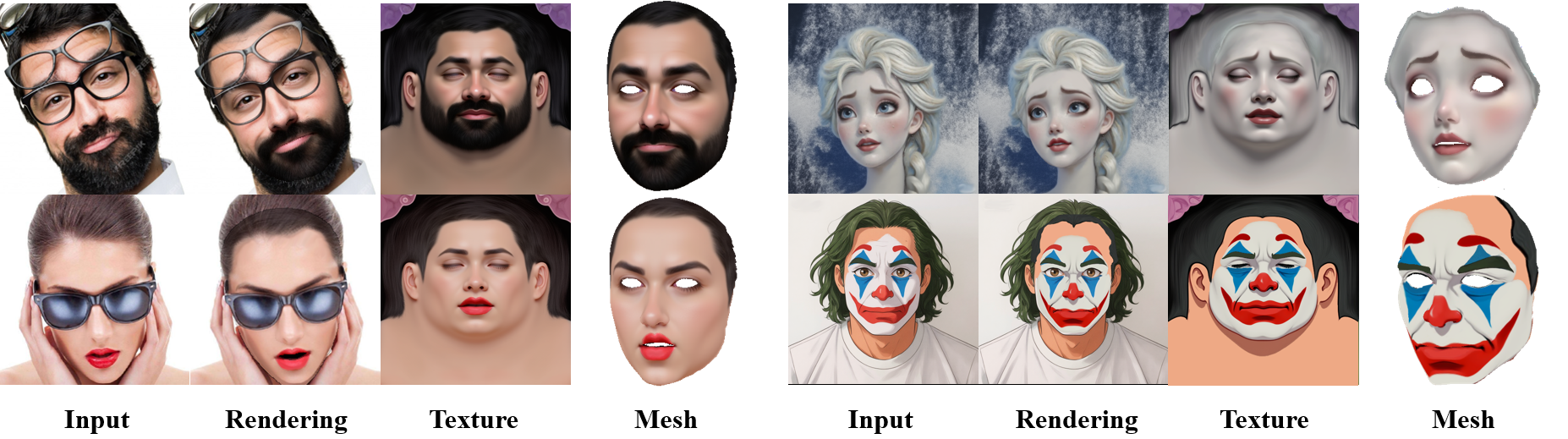}
    }
    \captionof{figure}{\methodName{} is an end-to-end framework capable of reconstructing high-fidelity and topologically consistent facial textures from a single input image within seconds. It demonstrates strong robustness against facial occlusions (e.g., hands, glasses and masks), while generalizing effectively across a wide range of stylized and in-the-wild facial images. As shown in Fig.~\ref{fig:style_trans} and Fig.~\ref{fig:abl21}, our method also enables subsequent texture editing, allowing convenient and coherent modifications. For better visual presentation, in the rendering results we additionally crop the glasses region from the input image and overlay it onto the rendered face.}
  \label{fig:teaser}
 \end{center}%
    }]
\begin{abstract}

We propose \methodName{}, an end-to-end diffusion-based framework for reconstructing high-quality and editable facial UV textures from multi-style facial images. Existing texture reconstruction methods face two major limitations: (1) Fragility due to reliance on 3D geometry priors, which are difficult to estimate accurately, especially under facial occlusions or in stylized domains; and (2) A lack of semantic disentanglement, inhibiting region-specific texture editing and style transfer. Our work addresses both challenges simultaneously.

Our core innovation is a geometry-free pipeline that directly maps a 2D face image to its corresponding editable UV texture. We introduce two key techniques: First, to address the challenge of UV misalignment common in diffusion generation, we introduce a gradient-guided refinement strategy at inference time, which explicitly corrects structural consistency. Second, we leverage the inherent semantic distribution capability of diffusion models and design a novel training paradigm to enhance this tendency, enabling semantic-aware editing of facial texture. Furthermore, to address the data scarcity in multi-style texture reconstruction, we construct \datasetName{}, the first comprehensive paired texture reconstruction dataset covering realistic and diverse stylized domains.

To the best of our knowledge, \methodName{} is the first geometry-free inference framework that achieves robust, style-consistent, and editable facial texture reconstruction across diverse domains. Our method achieves state-of-the-art performance on multiple facial texture benchmarks. Codes and the pretrained model weights will be released on \href{https://github.com/XZT24/OMGTex}{GitHub}.
\end{abstract}    
\section{Introduction}
\label{sec:intro}
3D face modeling and texture reconstruction are fundamental problems in computer vision. While significant progress has been made in reconstructing realistic facial geometry and textures from monocular images, generating editable and topologically consistent stylized textures remains a major challenge, despite its wide demand in applications such as gaming, virtual reality, and 3D animation.

Unlike real human faces, stylized avatars often exhibit exaggerated shapes, abstract shading, artistic brush strokes, and varying degrees of realism, which render traditional texture reconstruction methods ineffective. Beyond the data scarcity in the stylized facial domain, existing approaches face two fundamental limitations when applied to stylized inputs:
(1) Dependence on precise geometry priors. Accurate texture reconstruction typically assumes the availability of a topologically aligned 3D geometry which is then used for optimization or projecting partial textures. However, this assumption contradicts practical scenarios. In stylized domains, reconstructing precise and topology-consistent geometry is particularly difficult. In realistic domains, common facial occlusions such as hands and glasses also make it difficult to obtain accurate geometry. Moreover, even when accurate geometry is available, the resulting partial textures often incorrectly include artifacts from occluding objects.
(2) Lack of semantic disentanglement. Existing frameworks commonly learn the synthesis of the entire facial texture. This monolithic approach fundamentally inhibits intuitive, region-specific editing—such as adjusting skin tone, eyebrow shape, or lip appearance—and prevents seamless style transfer, both of which are commonly required for video game and VR avatars.

These limitations highlight the necessity of a new pipeline capable of generating multi-style, editable textures without relying on explicit geometry guidance. To this end, we present \methodName{}, an innovative texture reconstruction framework that reconstructs high-quality and editable textures from face images across diverse styles. Unlike prior methods that depend on accurate geometry fitting, via 3D Morphable Model (3DMM) \cite{blanz2023morphable} for example, followed by projection and diffusion-based completion, \methodName{} directly maps input facial images to editable UV textures using an improved end-to-end diffusion-based module.

Our key insight is that although diffusion models are powerful in generating diverse textures, they lack explicit UV constraints, leading to UV misalignment. This explains why prior diffusion-based texture works commonly rely on geometry fitting to provide partial UV maps. To overcome this challenge, we introduce a gradient-guided refinement strategy tailored for texture reconstruction, which corrects UV misalignment. Moreover, we observe that diffusion models inherently distribute different semantic facial features across different network blocks. We exploit this feature by further designing a novel training paradigm to enhance this tendency, enabling semantic-aware texture editing.

A key component of our framework is \datasetName{}, the first paired multi-style dataset of face images and corresponding textures. It contains 5,000 high-quality image–texture pairs and spans a broad spectrum of styles—including realistic, Japanese anime, Western comics, pixel art, sketch, oil-painted, Disney-style, and other stylized domains—and is constructed from both artist-created data and artist-refined synthetic samples. We further demonstrate that our framework can be extended to novel UV topologies. Further evidence for this extensibility is provided in the supplementary material.


By combining \datasetName{} with our reconstruction pipeline, \methodName{} achieves state-of-the-art performance on multiple benchmarks, excelling in reconstruction quality, style fidelity, efficiency, robustness, and UV consistency.

Our contributions can be summarized as follows:
\begin{itemize}
    \item We propose \methodName{}, a novel diffusion-based framework for facial texture reconstruction that requires no geometry priors, ensures UV structural consistency, robustness, and supports intuitive texture editing.
    \item We construct \datasetName{}, the first paired dataset of multi-style facial images and UV textures, which supports the training and evaluation of the facial texture reconstruction.
    \item We conduct comprehensive experiments on facial texture reconstruction from a single image input. We further apply our framework on regional texture editing and texture style transfer, showing the effectiveness of the proposed framework.
\end{itemize}

\section{Related Works}
\label{sec:related}

\vspace{-4mm}
\noindent \paragraph{3D Face Reconstruction}
Existing research on 3D face geometry reconstruction can be broadly categorized into two streams: reconstruction for realistic faces and reconstruction for stylized faces.

For realistic faces, early methods such as DECA \cite{feng2021learning} and Deep3DFace \cite{deng2019accurate} relied on minimizing facial landmark losses for geometry recovery. More recent approaches integrate facial segmentation as guidance: for instance, Part Re-projection Distance Loss (PRDL) \cite{wang20243d} reformulates segmentation maps into 2D point sets and optimizes their spatial distribution, thereby improving reconstruction accuracy.

In the domain of stylized faces, initial efforts focused on dataset construction. Qiu et al. \cite{qiu20213dcaricshop} introduced a caricature-style dataset containing around 2,000 samples. Building on this, Jung et al. \cite{jung2022deep} proposed a parametric model for exaggerated faces, which defines a learnable deformation space. More recent work \cite{jang2024toonify3d} targets toonified facial reconstruction, leveraging surface normal information within a 3D GAN–based framework to extract geometry.

Despite these advances, both realistic and stylized reconstruction methods typically rely on 3D Morphable Model (3DMM) \cite{blanz2023morphable} to define deformation spaces, while supervision often comes from pretrained facial landmark or segmentation detectors. This design introduces two key limitations: (1) the limited expressiveness of 3DMM constrains reconstruction to a narrow range of styles; and (2) dependence on pretrained detectors reduces robustness, especially under occlusions caused by hands, glasses, or other common objects.

Meanwhile, the rapid progress of image-to-3D generation has opened new possibilities. These methods adopt data-driven diffusion models to bridge the gap between 2D supervision and 3D generation. DreamFusion \cite{poole2022dreamfusion}, as a pioneering work, proposed the Score Distillation Sampling (SDS) loss, demonstrating that purely 2D supervision can suffice for 3D synthesis. With the release of large-scale 3D datasets \cite{slim20253dcompat++, objaverseXL, collins2022abo, fu20213d}, subsequent works have achieved remarkable quality and diversity. However, unlike traditional face reconstruction, which requires consistent topology across different inputs for geometry and texture, current 3D generation methods do not inherently guarantee these constraints. On the other hand, they exhibit superior robustness and generalization across diverse styles.

Motivated by these observations, our work leverages Hunyuan3D \cite{lai2025hunyuan3d} to obtain coarse geometry from reference images when constructing our dataset. We then deform a template mesh to fit the target shape, enabling the dataset to cover a wider range of facial styles while maintaining the consistent topology across different inputs.

\noindent \paragraph{Conditional Diffusion Model}

Diffusion models, exemplified by DDPM \cite{ho2020denoising}, achieve high-quality generation by gradually injecting Gaussian noise in the forward process and learning to denoise in the reverse process. Compared with GANs \cite{goodfellow2020generative}, they demonstrate superior sample diversity and training stability. Large-scale text-conditional diffusion models such as Imagen and Latent Diffusion / Stable Diffusion adopt a dominant paradigm: textual embeddings—produced by Transformers, CLIP, or large-scale language–vision encoders—are injected into the U-Net denoiser through cross-attention.

To enhance controllability over geometry and structure, a series of extensions have been proposed. ControlNet \cite{zhang2023adding}, T2I-Adapter \cite{mou2023t2i}, and GLIGEN \cite{li2023gligen} introduce strong geometric priors (e.g., edges, depth maps, surface normals, human poses, layouts, bounding boxes) by either freezing the backbone model, attaching lightweight adapters, or injecting regional conditions into the diffusion process. 

Our work is built upon an improved Diffusion Transformer (DiT) framework \cite{labs2025flux1kontextflowmatching, flux2024}. Unlike previous diffusion-based texture reconstruction approaches \cite{li2024uv, zhou2024ultravatar, yang2025freeuv} that rely on ControlNet to constrain texture structures, we adopt a novel gradient-guided strategy instead. The rationale behind this design choice is discussed in detail in Sec.\ref{sec:editable_uv}.

\begin{figure}
\centering
\includegraphics[width=0.98\linewidth]{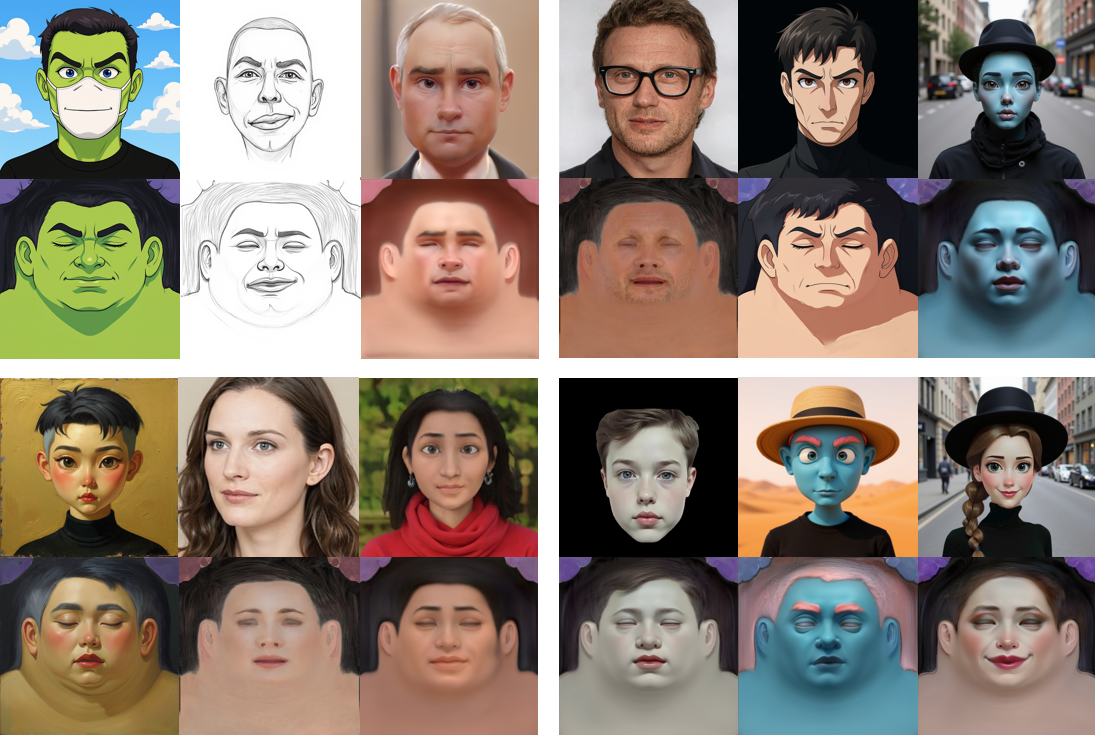}
\caption{The visualization of our \datasetName.}
\vspace{-3mm}
\label{fig:vis_texhub}
\end{figure}
\noindent \paragraph{Facial Texture Reconstruction}
OSTEC \cite{gecer2021ostec} is one of the earliest works on photorealistic facial texture reconstruction, while FaceRefiner \cite{li2024facerefiner} builds upon it with a post-optimization framework. FFHQ-UV \cite{bai2023ffhq} introduces a large-scale normalized UV texture dataset derived from FFHQ \cite{karras2019style}, along with a StyleGAN-based optimization scheme that improves 3DMM reconstruction quality and enhances rendering realism.
Early methods in this line primarily focus on photorealistic textures and typically rely on time-consuming post-optimization procedures.

With the rapid development of diffusion models, an increasing number of diffusion-based facial texture reconstruction frameworks have emerged in recent years. Since diffusion models inherently lack explicit structural constraints during training, UV-IDM \cite{li2024uv} first fits a 3DMM to the input image, projects it into a partial UV texture map, and uses it as a condition for a diffusion-based inpainting process, thereby ensuring UV structural consistency. Building on this, Ultr-Avatar \cite{zhou2024ultravatar} introduces an additional Diffuse Color Extraction Diffusion Model to explicitly decouple lighting effects from albedo. To address the shortage of stylized facial texture data, Soap \cite{liao2025soap} synthesizes multi-view images via diffusion and projects them into complete UV textures. Similarly, the state-of-the-art FreeUV \cite{yang2025freeuv} tackles the limited availability of texture data by introducing a ground-truth-free training framework that separates the learning process into two complementary components: appearance extraction and structural control. During training, only partial UV textures are required for appearance extraction, without the need for complete ground-truth facial textures.

Despite these advances, existing methods still face three key challenges:
(1) All current approaches \cite{li2024uv, zhou2024ultravatar, yang2025freeuv, liao2025soap, bai2023ffhq} rely on 3DMM geometry to extract either full or partial textures. This dependency limits their applicability to specific styles, constrains efficiency, and leads to severe performance degradation or outright failure when the geometric priors are unavailable or inaccurate.
(2) Partial-projection-based methods \cite{li2024uv, zhou2024ultravatar, yang2025freeuv} are prone to errors caused by facial occlusions, while fully projection-based approaches, such as Soap \cite{liao2025soap}, additionally suffer from multi-view inconsistency issues.
(3) Current frameworks are only capable of reconstructing textures within narrow stylistic domains, leaving the need for paired, high-quality, multi-style texture data still unmet.

In contrast, our framework constructs the paired multi-style texture dataset and bypasses the reliance on projected or partial UV maps. Through a novel training strategy, our method achieves strong robustness against stylized inputs and common occlusions, while simultaneously preserving texture editability. Furthermore, as demonstrated in the supplementary material, we show that with a small amount of additional data, \methodName{} can be extended to new UV topologies, further broadening its applicability.


\section{\datasetName: Multi-Style Facial Texture Dataset} 
\label{sec:dataset}

Existing facial texture reconstruction methods \cite{li2024uv, zhou2024ultravatar, yang2025freeuv, bai2023ffhq} are insufficient for handling multi-style inputs. A major challenge is the dearth of datasets for stylized facial images annotated with ground-truth facial UV-textures, which impedes not only the training of algorithms but also the evaluation of frameworks. To overcome this limitation, we adopt an optimization framework AvatarTex \cite{qiu2025avatartexhighfidelityfacialtexture} for dataset construction. 

For real facial texture, FFHQ-UV \cite{bai2023ffhq} proposed an efficient pipeline for data construction. To adapt this methodology to facial portraits of arbitrary styles, AvatarTex introduces two key modifications over FFHQ-UV to obtain a reasonable initial facial UV-texture: (1) For geometry, instead of relying on a 3D Morphable Model \cite{blanz2023morphable}, it first obtains the facial geometry $M$ using a general 3D generative model \cite{wu2024unique3d} and registers it to a canonical topology $M_v$ through Non-rigid Iterative Closest Point (NICP) alignment \cite{amberg2007optimal}, serving as geometry guidance to obtain the projected texture $T_{proj}$. (2) For UV-texture, it employs a fine-tuned texture inpainter diffusion network to complete the projected texture $T_{proj}$, producing $T_{init}$. The initialization $T_{init}$ serves as the starting point for StyleGAN-based \cite{Karras2019stylegan2} optimization, generating a high-quality texture $T$ for a stylized facial image. 

We first utilized AvatarTex \cite{qiu2025avatartexhighfidelityfacialtexture} to obtain a collection of paired facial images and corresponding UV textures. However, since AvatarTex is still a geometry-dependent framework built upon StyleGAN2 \cite{Karras2019stylegan2}, its expressive capability for diverse artistic styles and robustness to occlusion and positional variations remain limited — particularly for styles with difficult geometry (e.g., pixel art or caricatured cartoon styles).

To address these limitations, we subsequently employed FLUX.Kontext \cite{labs2025flux1kontextflowmatching} to augment the obtained image–texture pairs in terms of style diversity, as well as spatial and occlusion variations; implementation details can be found in the supplementary material. Finally, we invited experienced artists to further refine the augmented textures, ensuring that their structure and details are accurately aligned with the corresponding facial images.

Fig.~\ref{fig:vis_texhub} illustrates the overall composition and style diversity of our \datasetName{} dataset. This comprehensive dataset serves as both the training data for our framework and a crucial evaluation benchmark in our experiments.

\begin{figure}[h]
\centering
\includegraphics[width=0.98\linewidth]{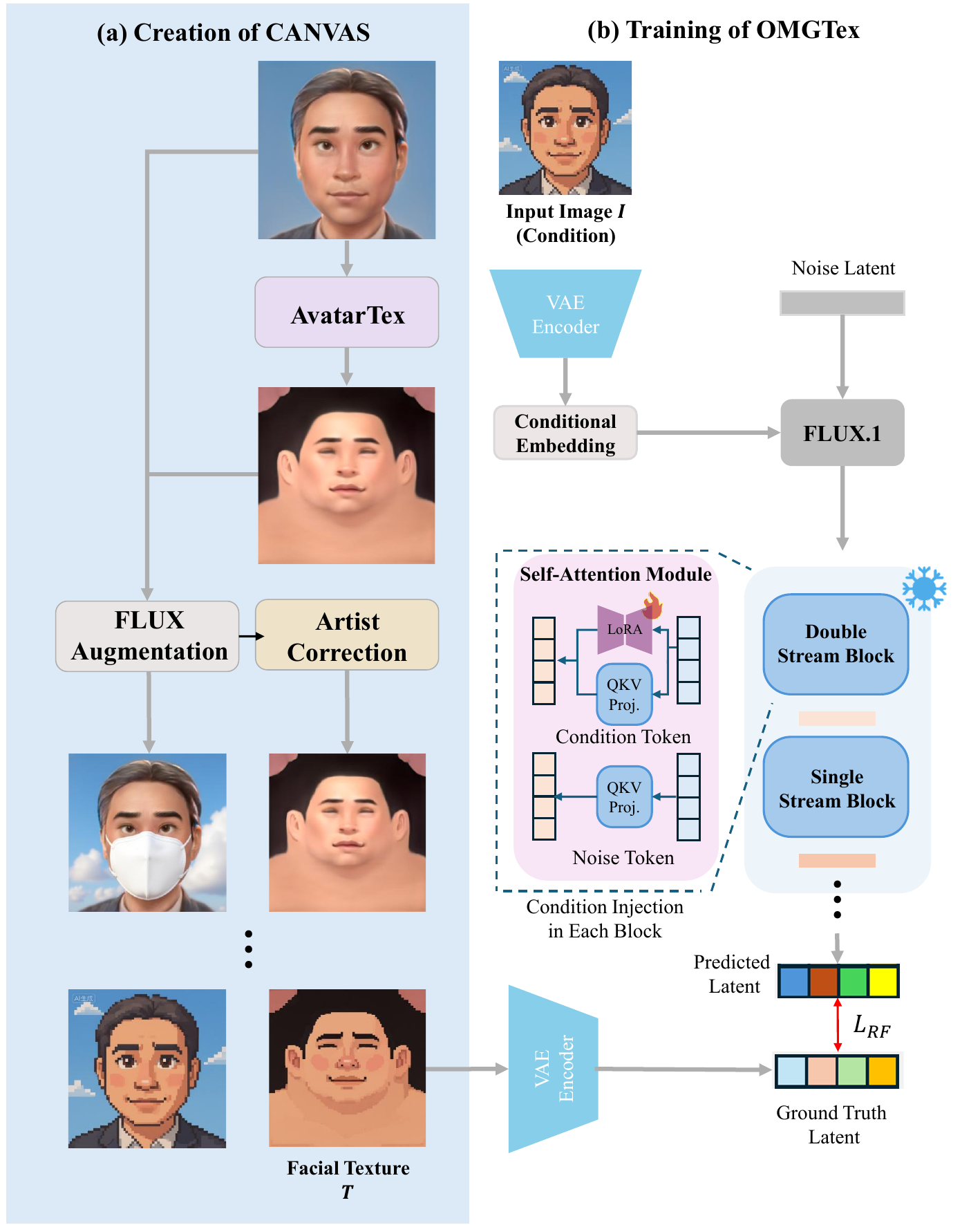}
\caption{The pipeline of \methodName, including (a) Creation of \datasetName{}, (b) Training of \methodName{}. }
\vspace{-3mm}
\label{fig:pipeline}
\end{figure}

\section{\methodName{}}
\label{sec:methods}



    



The primary goal of \methodName{} is to reconstruct high-quality facial textures $T$ from a single facial image $I$, without geometric guidance. To ensure our algorithm possesses sufficient generalization capability for arbitrary input faces, we adopt the Diffusion Transformer (DiT)-based FLUX.1.DEV \cite{flux2024} as our reconstruction backbone.

As illustrated in Fig.\ref{fig:pipeline}, we improve a conditional image generation framework \cite{zhang2025easycontrol} to train a Low-Rank Adaptation (LoRA) \cite{hu2022lora} condition module. This trained LoRA enables the underlying FLUX network to generate the UV texture map corresponding to the conditional input face image. We optimize the training process with the flow-matching loss function \cite{lipman2022flow} as follows:
\begin{equation}
    L_{RF} = \mathbb{E}_{t, \epsilon \sim \mathcal{N}(0,I)} \Big[ \big\| v_\theta(z,t,c_i) - (\epsilon - x_0) \big\|^2 \Big].
\end{equation}

While the generated texture image is high-quality and robust to various input facial images, the layout of the result could not align with the specific UV-texture.

\paragraph{\textbf{Gradient-Guided Texture Alignment}}

\label{sec:editable_uv}

To ensure accurate texture structures, previous works \cite{li2024uv, zhou2024ultravatar, yang2025freeuv} typically use incomplete textures obtained via projection as conditional inputs, which provide strong structural priors for guiding texture generation. In contrast, we directly use facial images as conditions, which lack explicit UV structural constraints. Specifically, we observe two types of structural inconsistencies in the reconstructed textures as shown in Fig.~\ref{fig:abl1}: (1) the eyes sometimes appear open when they should remain closed, and (2) the facial components become vertically misaligned. A naive solution would be to inject additional structural conditions through ControlNet \cite{zhang2023adding}. However, in our experiments, we observed that the reconstructed textures still exhibit deviations from the canonical topology. Meanwhile, the control signals from ControlNet may conflict with existing conditions, resulting in a degradation of reconstruction quality. Motivated by this observation, we investigate the underlying causes and explore potential solutions.

Our exploration begins with a frequency-based classification of image information. For instance, in textures, fine details such as skin wrinkles or eyebrow shapes correspond to high-frequency components. In contrast, broader structures, such as skin tone or general texture layout, correspond to low-frequency components. The diffusion process can be expressed as the following ordinary differential equation (ODE):
\begin{equation}
\frac{dx_t}{dt} = f(x_t, t), \quad x_t = (1 - t)x_0 + t x_1,
\end{equation}
where $x_0$ denotes the clean data sample, $x_1$ is the noise endpoint, and $x_t$ represents an intermediate point along the deterministic trajectory governed by the flow field $f(x_t, t)$.

Applying a Fourier transform, the signal-to-noise ratio (SNR) of a frequency component $\omega$ at timestep $t$ can be written as:
\begin{equation}
\text{SNR}(\omega, t)
= \frac{(1 - t)^2 \, \left|\hat{x_0}(\omega)\right|^2}{t^2 \, \left|x_1(\omega)\right|^2}, 
\end{equation}
where $\hat{x}_0(\omega)$ denotes the Fourier coefficient of $x_0$ at frequency $\omega$. Since $x_1$ is the noise end point, the SNR degradation depends solely on the spectral energy distribution of the original signal $x_0$.

Natural images typically exhibit low-pass characteristics, which can be approximately modeled as:
\begin{equation}
|\hat{x}_0(\omega)|^2 \propto |\omega|^{-\alpha}, \quad \alpha > 0,
\end{equation}
where $\alpha$ is the attenuation coefficient. This implies that high-frequency components possess significantly lower energy compared to low-frequency components. Consequently, during diffusion, the SNR of high-frequency components decays much faster than that of low-frequency components.

This observation reveals an important insight: when training a diffusion model to reconstruct $x_0$ from $x_t$, the supervision signal is naturally stronger for high-frequency components. As a result, while high-frequency details are quickly corrupted during the forward diffusion process, they remain more responsive to guidance during denoising. In contrast, low-frequency structures are relatively insensitive to control signals, making them difficult to manipulate precisely—a phenomenon consistent with intuition.

Inspired by this insight, we hypothesize that simply injecting structural conditions via ControlNet does not guarantee precise control over texture structures. Moreover, additional structural constraints may conflict with existing conditions, potentially degrading reconstruction quality. Therefore, rather than injecting structural conditions, it is more reasonable to guide the denoising process directly through gradients with explicit UV-structure constraints.

To this end, we design an energy function that explicitly measures structural deviations. Concretely, we train a landmark detector $l(\cdot)$ tailored for facial textures, which predicts the landmark positions on the reconstructed texture $\hat{x}_0$. These predicted landmarks are compared against canonical landmark positions $l^\ast$, and the discrepancy is quantified using an $\ell_2$ loss:
\begin{equation}
E(\hat{x}_t) = \|\, l(\hat{x}_t) - l^\ast \,\|_2^2.
\end{equation}

Based on this formulation, we propose an iterative gradient-guided procedure during inference, which is inspired by classifier guidance \cite{ho2020denoising}:
\begin{equation}
\tilde{x}_t = \hat{x}_t - \eta \nabla_{\hat{x}_t} E(\hat{x}_t),
\end{equation}
where $\eta$ is a step size controlling the influence of the structural constraint, and $\nabla_{\hat{x}_t} E(\hat{x}_t)$ denotes the gradient of the energy function with respect to the current prediction.

\begin{figure}[h]
\centering
\includegraphics[width=0.98\linewidth]{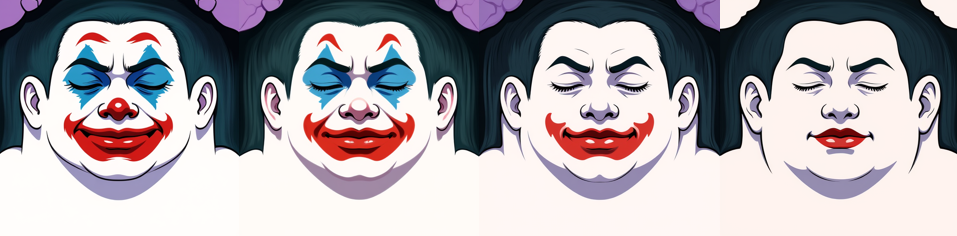}
\caption{Visualization of the degradation of the texture.}
\vspace{-3mm}
\label{fig:inter}
\end{figure}
\section{Application: Facial UV-Texture Editing Based on \methodName{}}
We demonstrate that our \methodName{} enables semantic-aware editing for facial texture. To this end, we conduct ablation experiments across different layers of the Transformer blocks in \methodName{}. 
Specifically, we ablate the feature extraction capability of a given attention layer by scaling its attention weight matrix with a small constant $\varepsilon$:

\begin{equation}
\tilde{\mathbf{A}}^{(l)} = \varepsilon \cdot \mathbf{A}^{(l)}, \quad \varepsilon \ll 1,
\end{equation}

where $\mathbf{A}^{(l)}$ denotes the attention weight matrix at layer $l$. Fig.~\ref{fig:inter} illustrates the progressive degradation of texture features as more layers are ablated.

We discover that the double stream blocks and the single stream blocks (consistent with the definition in \cite{flux2024}) in our trained \methodName{} tend to learn different properties of the condition input facial image. The double stream blocks determine the visual style of the generated facial texture, while the single stream blocks learn to generate the identity feature of the condition input. Furthermore, the degradation in the single stream blocks follows a certain rule: in this case, features around the nose degrade first, followed by the eyes, and finally the mouth. This suggests that each attention layer in the single stream blocks of FLUX contributes differently to distinct texture components.

To exploit this observation, we implement two applications that explicitly reinforce semantic texture editing.

\paragraph{Texture Style Transfer:}
Given an identity condition $I_{id}$ and a style condition $I_{st}$, we first generate the attention output features $F_{id}$ of $I_{id}$ with \methodName{}, including the double stream block features $F_{id}^{double}$ and the single stream block features $F_{id}^{single}$ . In the reconstruction process of $I_{s}$, we replace the corresponding attention output in single stream blocks $F_{st}^{single}$ with $F_{id}^{single}$, achieving the style transfer for facial texture.

\begin{figure}
\centering
\includegraphics[width=0.98\linewidth]{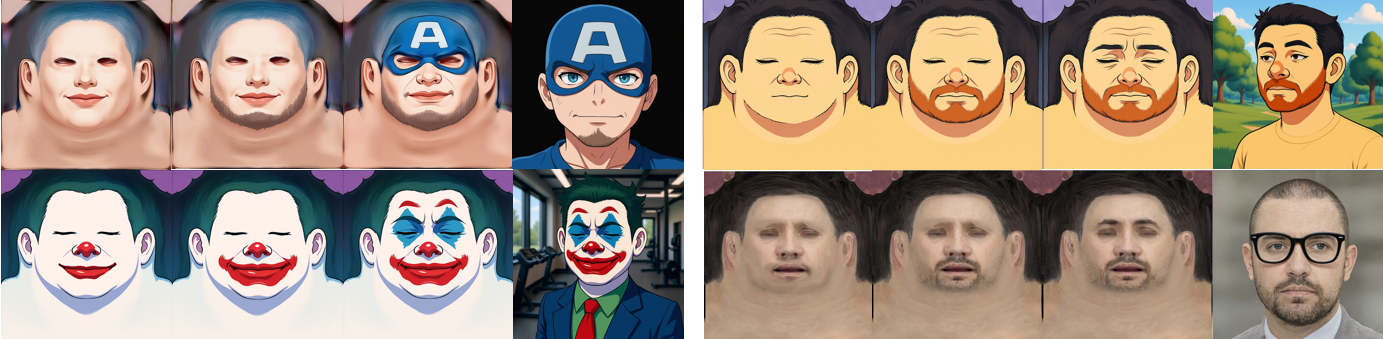}
\caption{To facilitate supervised training for editable texture generation, we construct a dedicated dataset derived from the post-processed outputs of \datasetName{}. }
\vspace{-3mm}
\label{fig:edit_dataset}
\end{figure}
\begin{figure}[h]
\centering
\includegraphics[width=0.98\linewidth]{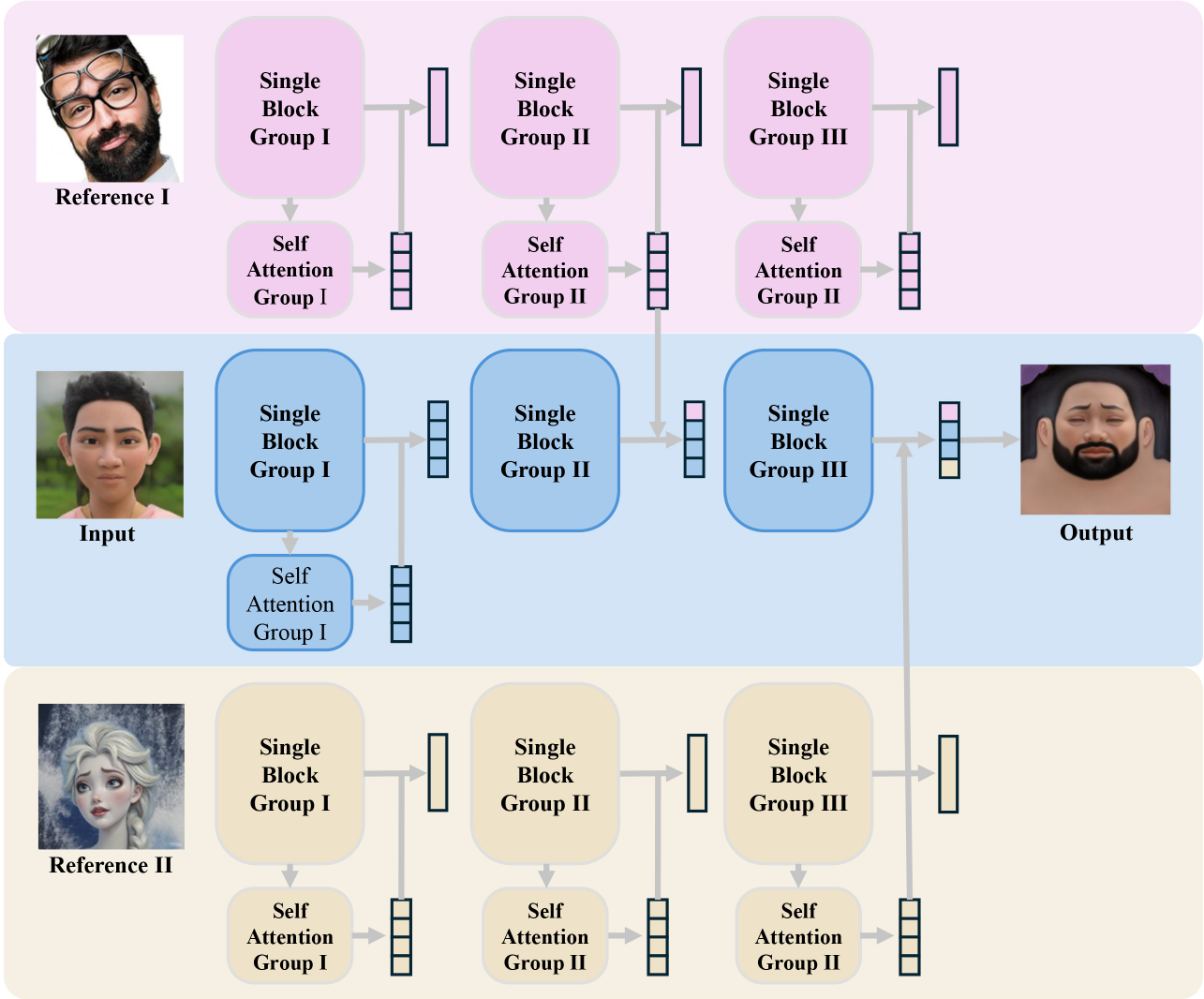}
\caption{The illustration of the regional editing of our \methodName. }
\vspace{-3mm}
\label{fig:pipeline_edit}
\end{figure}
\paragraph{Regional Texture Editing:}
Given a source identity image $I_{A}$, we aim to modify the specific region into the texture of a condition image $I_{B}$. For brief illustration, we decompose each texture into three regions: the overall skin tone and fixed structural components; the mouth region along with its surrounding details; the eyebrows. To enhance the accuracy of editing, we slightly modify the train strategy of our framework to enhance semantic disentanglement across attention blocks.

We first augment the facial texture image in CANVAS through a layer-wise editing process, as shown in Fig.~\ref{fig:edit_dataset}. Similar to the construction of our dataset, we then divide the attention blocks into three functional groups as shown in Fig.~\ref{fig:pipeline_edit}: the first group is responsible for generating the overall skin tone and coarse texture structure; the second group focuses on the mouth and its surrounding features; and the third group captures the eyebrows and their surrounding regions. During training, whenever the model reaches the boundary between two groups of attention layers, it ablates all subsequent layers with a fixed probability $p$, and directly outputs the intermediate features. These features are then supervised by the corresponding augmented partial textures. By explicitly attributing different semantic information across different Transformer blocks, our model enables fine-grained local feature editing. As shown in Fig.~\ref{fig:abl21}, given an Input and Reference, the Input provides the base skin tone information, while Reference contributes the mouth-region details or eyebrow features. 

We demonstrate the effectiveness of our application in Sec.~\ref{sec:app_show}.

\section{Experiments}
\label{sec:exps}

\begin{figure*}
\centering
\includegraphics[width=0.95\linewidth]{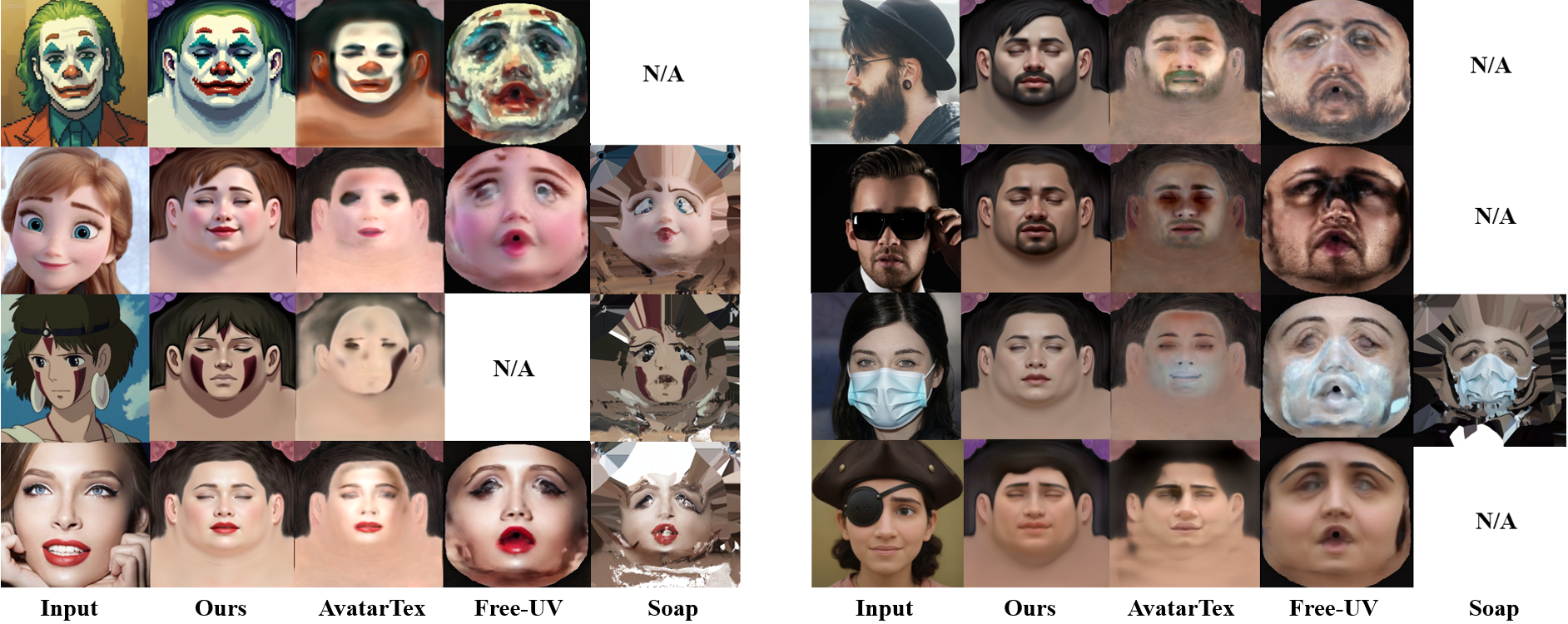}
\vspace{-3mm}
\caption{Qualitative Comparison. The left column demonstrates the robustness of our \methodName{} to stylized inputs, while the right column highlights its robustness to facial occlusions and head pose. "N/A" indicates that the input could not be fitted by the 3DMM or no eyes are detected, leading to a failure in texture reconstruction.}

\label{fig:comp}
\end{figure*}

\begin{table*}[htb]
\centering
\resizebox{\textwidth}{!}{%
\begin{tabular}{c|cccc|cccc|cccc}
\multirow{2}{*}{Method} 
& \multicolumn{4}{c|}{\textbf{FFHQ}} 
& \multicolumn{4}{c|}{\textbf{LPFF}} 
& \multicolumn{4}{c}{\textbf{CANVAS}} \\ \cline{2-13}
& PSNR$\uparrow$ & SSIM$\uparrow$ & LPIPS$\downarrow$ & FID$\downarrow$
& PSNR$\uparrow$ & SSIM$\uparrow$ & LPIPS$\downarrow$ & FID$\downarrow$
& PSNR$\uparrow$ & SSIM$\uparrow$ & LPIPS$\downarrow$ & FID$\downarrow$ \\ \hline
AvatarTex  & \textbf{30.03} & \textbf{0.85} & \textbf{0.16} & \textbf{22.15}
           & \underline{27.91} & \underline{0.79} & \underline{0.25} & \underline{38.93}
           & 23.93 & 0.74 & 0.34 & 60.02 \\
Soap       & 24.45 & 0.74 & 0.31 & 50.97
           & 22.76 & 0.70 & 0.35 & 60.71
           & 21.28 & 0.68 & 0.42 & 68.63 \\
Free-UV    & 29.18 & 0.80 & 0.22 & 35.36
           & 26.12 & 0.77 & 0.27 & 45.91
           & \underline{24.46} & \underline{0.75} & \underline{0.31} & \underline{50.89} \\
\textbf{Ours} 
           & \underline{29.75} & \underline{0.82} & \underline{0.18} & \underline{28.12}
           & \textbf{28.92} & \textbf{0.80} & \textbf{0.23} & \textbf{36.78}
           & \textbf{27.22} & \textbf{0.77} & \textbf{0.26} & \textbf{43.55} \\
\end{tabular}%
}
\caption{Quantitative comparison across FFHQ, LPFF, and CANVAS datasets. Our method consistently achieves leading performance among inference-based frameworks across all datasets. However, on the FFHQ dataset, our results are lower than the optimization-based AvatarTex. This is primarily because, when accurate geometry is available, optimization-based approaches can produce more precise and fine-grained reconstructions, albeit at the cost of geometric dependence and significantly longer runtimes.}
\label{tab:rebuttal}
\end{table*}

\begin{table}[htb]\small

\centering
\begin{tabular}{c|cccc}
 & AvatarTex & Soap & Free-UV & \textbf{Ours} \\ \hline
Time (s) & 90 & 360 & 20 & \textbf{7} \\
\end{tabular}
\caption{Inference time comparison across different methods.}
\label{tab:time}
\end{table}

\subsection{Comparison on Facial Texture Reconstruction}

\paragraph{Qualitative Comparison}
We compare our method on in-the-wild inputs against several state-of-the-art approaches in texture reconstruction, including Free-UV \cite{yang2025freeuv} for realistic-domain faces and AvatarTex \cite{qiu2025avatartexhighfidelityfacialtexture} and SOAP \cite{liao2025soap} for stylized domains.

Experimental results demonstrate that Free-UV, trained primarily on real-world facial data, struggles to generalize to stylized domains. The multi-view projection–based SOAP suffers from severe artifacts caused by facial occlusions and inconsistencies across views. Moreover, all three methods rely heavily on accurate geometric priors, making them vulnerable when such priors are unreliable or when the input image contains occlusions—leading to blurred or misaligned textures in the texture initialization of AvatarTex, or failed reconstructions and the inclusion of occluders in Free-UV and SOAP.

\paragraph{Quantitative Comparison}
We perform a quantitative comparison of our approach with existing state-of-the-art methods in rendering space across three distinct datasets.
For evaluation on realistic face inputs, we randomly select 1,000 facial images from FFHQ ~\cite{karras2019style} as the test set. To further evaluate the robustness of our method to large pose and positional variations, we additionally sample 1,000 images from the Large Pose Face Dataset (LPFF) \cite{Wu_2023_ICCV}.
For stylized face inputs, we conduct evaluations on our \datasetName{} dataset, utilizing 500 dedicated test samples. To ensure a fair comparison, we transform all reconstruction results into the canonical UV layout of the FLAME model and adopt the same geometry as FreeUV.

As shown in Tab.~\ref{tab:rebuttal}, our inference-based approach achieves superior performance on the FFHQ dataset, placing second only to the optimization-based AvatarTex \cite{qiu2025avatartexhighfidelityfacialtexture}. More importantly, on the more challenging LPFF and \datasetName{} datasets, our method consistently outperforms all competitors by a significant margin, clearly demonstrating the advantages of our geometry-free design in handling diverse styles, occlusions, and large-pose inputs.

Tab.~\ref{tab:time} compares the inference time. AvatarTex requires iterative refinement of the latent code of StyleGAN, while SOAP further necessitates multi-view image generation. Free-UV is capable of direct texture reconstruction via inference; however, it still requires additional time to reconstruct the underlying face geometry guiding their texture reconstruction process. In contrast, our method bypasses geometric priors entirely and reconstructs textures in a single, end-to-end diffusion step, achieving a significant advantage in reconstruction efficiency.

\begin{figure}
\centering
\includegraphics[width=0.95\linewidth]{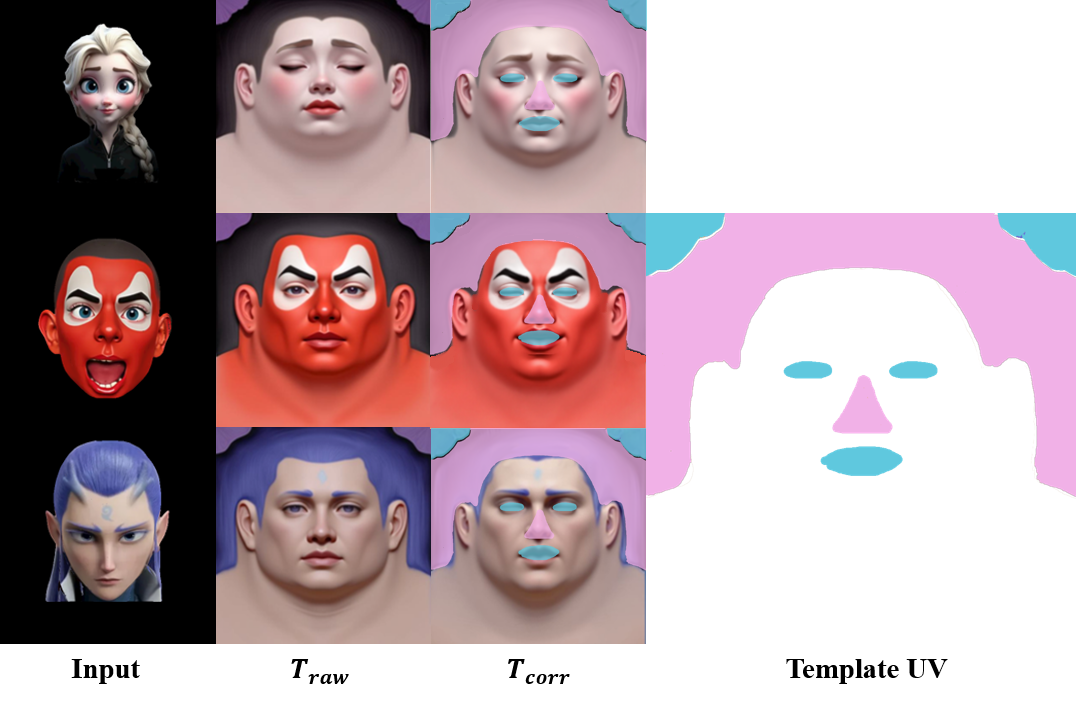}
\caption{Visualization on the effectiveness of Gradient-Guided Texture Alignment. $T_{raw}$: the directly outputs; $T_{corr}$: our results.}
\vspace{-3mm}
\label{fig:abl1}
\end{figure}


\begin{table}[htb]
\centering
\resizebox{\linewidth}{!}{
\begin{tabular}{c|ccccc}
Method & PSNR$\uparrow$ & SSIM$\uparrow$ & LPIPS$\downarrow$ & FID$\downarrow$ & L2$\downarrow$ \\ \hline
\textbf{Ours}                & \textbf{27.22}  & \textbf{0.77}  & \textbf{0.26}  & \textbf{43.55} &  \textbf{1.35}  \\
w/o Grad Guide & 25.16  & 0.73  & 0.30  & 50.19 &  2.98  \\
ControlNet & 25.93  & 0.75  & 0.29  & 48.64 &  2.64  \\
\end{tabular}}
\vspace{-2mm}
\caption{The ablation results on the testset of our CANVAS.}
\vspace{-2mm}
\label{tab:control}
\end{table}

\subsection{Ablation Study}
We conduct comprehensive ablation studies to evaluate the individual effectiveness of our UV-Consistent Texture Reconstruction components. As shown in Fig.~\ref{fig:abl1}, the baseline reconstruction without our strategy exhibits clear UV misalignment, as discussed in Sec.~\ref{sec:editable_uv}.  These issues manifest in the Result $T_{raw}$ column, where the reconstructed textures show inconsistent facial component alignment and artifacts such as opened eyes. By incorporating our UV-consistent reconstruction strategy, the texture quality and structural alignment improve significantly, as illustrated in Result $T_{corr}$. Tab.~\ref{tab:control} further compares the results of the proposed gradient-guided strategy with directly applying ControlNet \cite{zhang2023adding}. The results demonstrate that while ControlNet can partially correct missing texture structures, its effectiveness is inferior to that of the dedicated gradient-guided approach. Moreover, conflicts among different control signals often lead to degradation in overall reconstruction quality.

\begin{figure}
\centering
\includegraphics[width=0.90\linewidth]{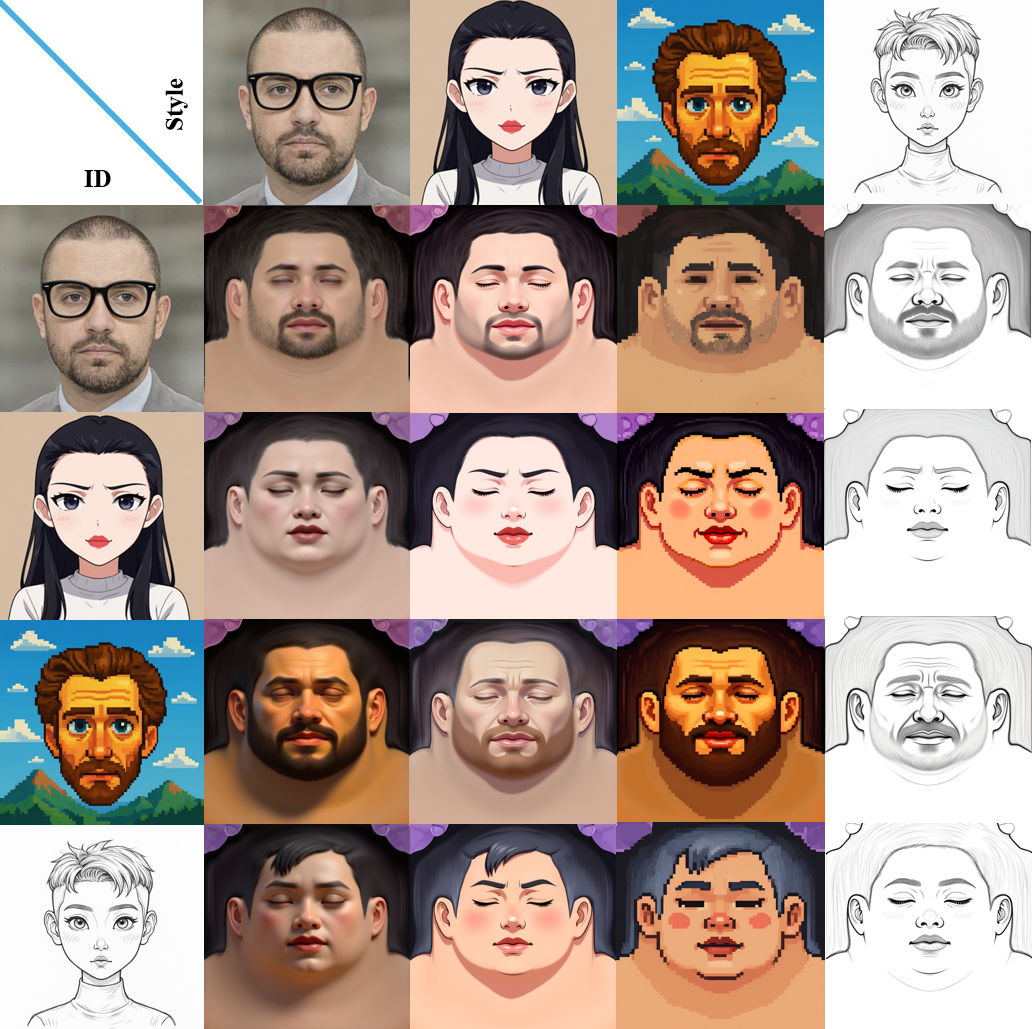}
\caption{The visualization of facial texture style transfer.}
\vspace{-3mm}
\label{fig:style_trans}
\end{figure}
\vspace{-1mm}
\begin{figure}
\centering
\includegraphics[width=0.98\linewidth]{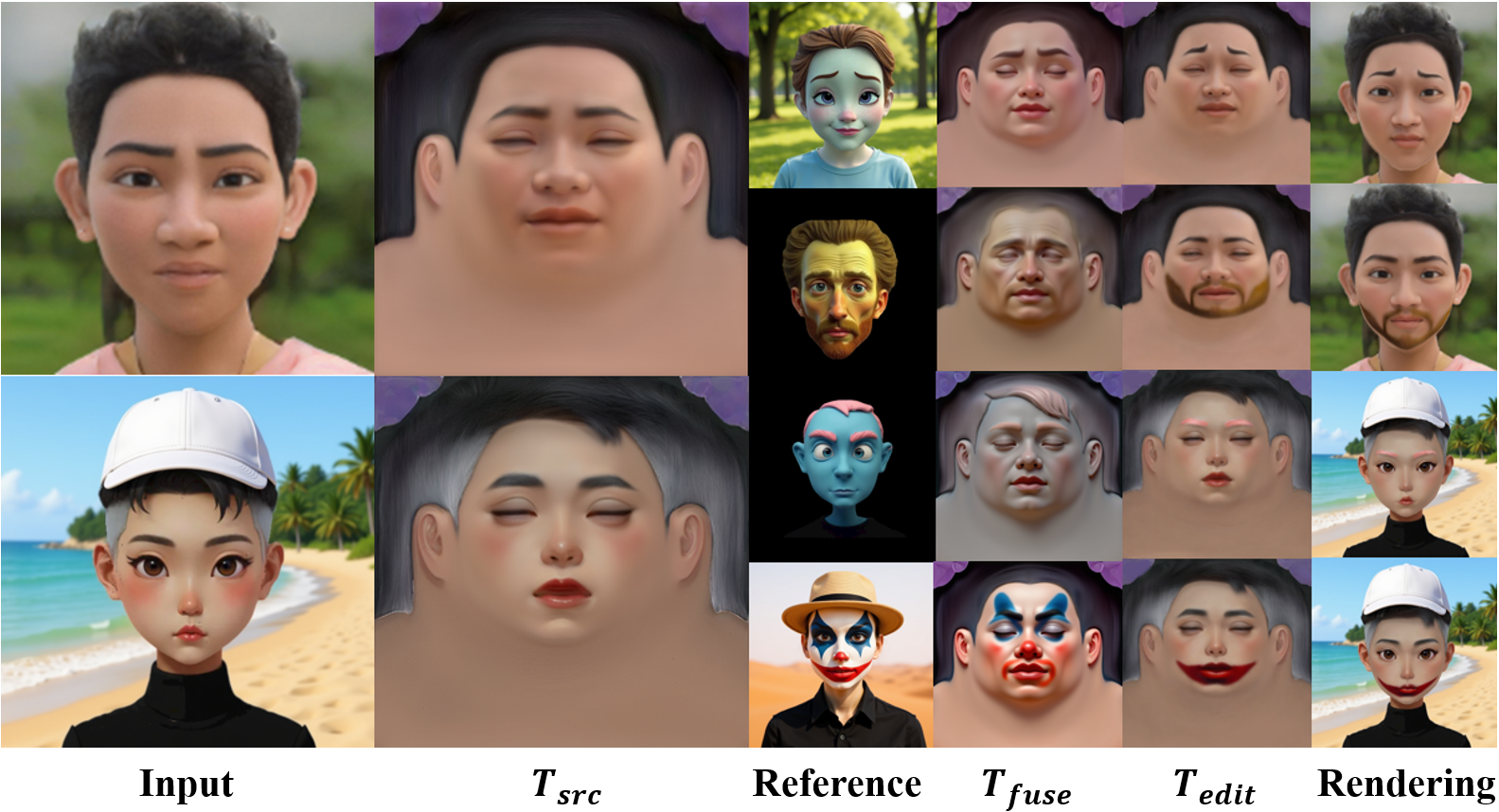}
\caption{Visualization of regional texture editing (for eyebrow and around the mouth respectively). $T_{src}$: the source texture; $T_{fuse}$: the result obtained by directly blending features from the reference. $T_{edit}$: the edited texture using our proposed \methodName.}
\vspace{-3mm}
\label{fig:abl21}
\end{figure}


\subsection{Application}
\label{sec:app_show}

We show visualization results of texture style transfer and regional texture editing with our \methodName{}.

As Fig.~\ref{fig:style_trans} shown, given an identity image (the first column) and a style condition (the first row), \methodName{} could generate the facial texture with combination of target identity and visual style.

Fig.~\ref{fig:abl21} demonstrates editing results of the eyebrow region, the mouth region, and both regions simultaneously. In each case, $T_{src}$ represents the reconstruction without feature blending; $T_{edit}$ shows the result of semantic-aware editing using our proposed training strategy, where features from the reference image are fused into the input only at designated semantic regions. In contrast, $T_{fuse}$, which applies direct feature blending without semantic disentanglement, results in widespread changes across the entire texture, failing to achieve localized editing.
\section{Conclusion}
We present \methodName{}, an end-to-end framework for facial texture reconstruction without geometric guidance. Our approach demonstrates strong robustness to both stylized inputs and facial occlusions, addressing a key limitation of diffusion-based methods.


\paragraph{Acknowledgements.}
The work was supported in part by Guangdong S\&T Programme with Grant No. 2024B0101030002, the Basic Research Project No. HZQB-KCZYZ-2021067 of Hetao Shenzhen-HK S\&T Cooperation Zone, by Guangdong Provincial Outstanding Youth Fund with No. 2023B1515020055, the Shenzhen Outstanding Talents Training Fund 202002, the NSFC with Grant No. 62293482, the Guangdong Research Projects No. 2017ZT07X152 and No. 2019CX01X104, the Guangdong Provincial Key Laboratory of Future Networks of Intelligence (Grant No. 2022B1212010001), and the Shenzhen Key Laboratory of Big Data and Artificial Intelligence (Grant No. SYSPG20241211173853027), the Guangdong Province Radio Science Data Center.

{
    \small
    \bibliographystyle{ieeenat_fullname}
    \bibliography{main}
}

\end{document}